\documentclass[runningheads]{llncs}
\usepackage{cite}
\usepackage{amsmath,amssymb}
\usepackage{mathtools}
\usepackage{xcolor}
\usepackage{xspace}
\usepackage{wrapfig}
\usepackage{graphicx}
\usepackage{tikz}
\usepackage{nicefrac}
\usepackage{todonotes}
\usepackage{wasysym}
\usepackage{tikz}
\usepackage{svg}
\usepackage[labelformat=parens,labelsep=quad, skip=3pt]{caption}
\usepackage{subcaption}
\usepackage{hyperref}
\usepackage{graphicx}
\usepackage[mode=buildnew]{standalone}
\usepackage[]{adjustbox}
\usepackage{lineno}
\usepackage{xcolor}

\usetikzlibrary{shapes,shadows,arrows,positioning,angles,quotes}
\tikzset{My Arrow Style/.style={single arrow, fill=black!15, anchor=base, align=center,text width=2.3cm}}
\tikzstyle{arrow} = [thick,->,>=stealth]
\usetikzlibrary{calc,trees,positioning,arrows,fit,shapes,calc}

\tikzstyle{startstop} = [rectangle, rounded corners, minimum width=2.2cm, minimum height=0.5cm,text centered, draw=black, fill=red!30]
\tikzstyle{io} = [trapezium, trapezium left angle=70, trapezium right angle=110, minimum width=1cm, minimum height=0.5cm, text centered, draw=black, fill=blue!30]

\tikzstyle{process} = [rectangle, minimum width=2.2cm, minimum height=0.5cm, text centered, draw=black, fill=orange!30]
\tikzstyle{decision} = [diamond, minimum width=1.5cm, minimum height=0.5cm, text centered, draw=black, fill=green!30]

\tikzstyle{arrow} = [thick,->,>=stealth]

\renewcommand{\paragraph}[1]{\smallskip\noindent\emph{#1}}

\newcommand*\pfatchAmsMathEnvironmentForLineno[1]{%
  \expandafter\let\csname old#1\expandafter\endcsname\csname #1\endcsname
  \expandafter\let\csname oldend#1\expandafter\endcsname\csname end#1\endcsname
  \renewenvironment{#1}%
     {\linenomath\csname old#1\endcsname}%
     {\csname oldend#1\endcsname\endlinenomath}}%

\let\doendproof\endproof
\renewcommand\endproof{~\hfill$\qed$\doendproof}

\usepackage{array,booktabs}
\newcolumntype{C}{>{$\displaystyle}c<{$}}
\usepackage[normalem]{ulem}
\useunder{\uline}{\ul}{}

\makeatletter

\let\epsilon\varepsilon
\let\phi\varphi

\makeatother

\begin{document}

\title{COOL-MC: A Comprehensive Tool for Reinforcement Learning and Model Checking\thanks{Download it from \url{https://zenodo.org/record/6948517\#.Yvff-OxBxhE}}
}

\titlerunning{COOL-MC: A Tool for RL and model checking}
%
\author{Dennis Gross\inst{1}
\and
Nils Jansen\inst{1}
\and
Sebastian Junges\inst{1}
\and
Guillermo A. P{\'e}rez\inst{2}
}

\authorrunning{D.\ Gross, N.\ Jansen, S.\ Junges, G.\ A.\ P{\'e}rez}
%
\institute{Radboud University, The Netherlands
\and
University of Antwerp -- Flanders Make, Belgium
}

\maketitle              

\begin{abstract}
This paper presents COOL-MC, a tool that integrates state-of-the-art reinforcement learning (RL) and model checking.
Specifically, the tool builds upon the OpenAI gym and the probabilistic model checker Storm.
COOL-MC provides the following features: (1) a simulator to train RL policies in the OpenAI gym for Markov decision processes (MDPs) that are defined as input for Storm, (2) a new model builder for Storm, which uses callback functions to verify (neural network) RL policies, (3) formal abstractions that relate models and policies specified in OpenAI gym or Storm, and (4) algorithms to obtain bounds on the performance of so-called permissive policies.
We describe the components and architecture of COOL-MC and demonstrate its features on multiple benchmark environments.
\end{abstract}

%
%
%

\section{Introduction}


Deep Reinforcement learning (RL) has created a seismic shift in how we think about building 
agents for dynamic systems~\cite{mnih2015human,levine2016end}.
It has triggered applications in critical domains like energy, transportation, and defense~\cite{nakabi2021deep,farazi2021deep,boron2020developing}.
An \emph{RL agent} aims to learn a near-optimal policy regarding some fixed objective by taking actions and perceiving feedback signals, usually rewards and observations of the \emph{environment}~\cite{sutton2018reinforcement}.
Unfortunately, learned policies come with no guarantee to avoid \emph{unsafe behavior} \cite{DBLP:journals/jmlr/GarciaF15}.
Generally, rewards lack the expressiveness to encode complex safety requirements \cite{DBLP:journals/aamas/VamplewSKRRRHHM22}
and it is hard to assess whether the training at a certain point in time is sufficient.

To resolve the issues mentioned above, verification methods like model checking~\cite{baier2008principles,DBLP:reference/mc/2018} are used to reason about the safety of RL, see for instance \cite{yuwangPCTL,DBLP:conf/formats/HasanbeigKA20,DBLP:conf/atva/BrazdilCCFKKPU14,DBLP:conf/tacas/HahnPSSTW19}.
However, despite the progress in combining these research areas, there is---to the best of our knowledge---no mature tool support that tightly integrates exact model checking of state-of-the-art deep RL policies.
One of the reasons is that policies obtained on an OpenAI environment may be incompatible with a related formal model (that can be used for model checking), and vice versa. These incompatibilities are often due to differences in their state or action spaces, or even their rewards. 
Another challenge is that verifying deep RL policies currently requires
different algorithms, data structures, and abstractions depending on the
architecture and size of a neural network (NN).


We present COOL-MC,
an open-source tool that integrates the OpenAI gym
with the probabilistic model checker Storm~\cite{dehnert2017storm}.
The purpose of the tool is \emph{to enable training RL policies while being able to verify them at any stage of the training process}. Concretely, we focus on supporting the following methodology as main use case for COOL-MC. An RL expert is tasked with training an RL policy with formal guarantees. These guarantees are captured as formal specifications that hold on a well-defined formal model, formulated by the expert. 
At any stage of the learning, the RL expert wants to establish whether the policy  has the desired properties as given by the formal model and specification.

\label{inputs} Thus, the input of COOL-MC consists of two models of the environment: (1) an \emph{OpenAI-gym compatible environment}, to train
an RL policy; (2) an \emph{Markov decision process (MDP)}, specified using the PRISM
language~\cite{kwiatkowska2004prism}, to verify the
policy together with a  formal specification e.g., a
probabilistic computation tree logic (PCTL)
formula. 
Only the MDP model of the environment is required: If no OpenAI-gym environment is given, COOL-MC provides a wrapper to cast
the MDP as an OpenAI gym environment.
The latter is done using a \emph{syntax-based simulator} that avoids
building the MDP explicitly in Storm.
Training of an RL policy can be done using any RL agent available in
the OpenAI gym. Any (trained) policy can then be formally verified via
Storm
using \emph{callback functions} that query the NN and
build the induced discrete-time Markov chain (DTMC) incrementally~\cite{DBLP:conf/concur/CassezDFLL05,DBLP:conf/tacas/DavidJLMT15}.
COOL-MC assumes no formal relation between the two given environment models. Of course, for the purpose of policy verification, the states of both have to be in some of relation. 
For this purpose, we support abstraction of models and policies. In particular, 
we employ a feature-based representation for MDPs~\cite{strehl2007efficient}.
Such features may, for instance, refer to concrete positions of agents in their environment, or to the fuel level of a car.
We offer the user the option to \emph{remap feature values}
and to define
\emph{abstractions} of their feature domains \cite{DBLP:conf/atva/JaegerJLLST19} (see Section~\ref{sec:ab}). This yields the opportunity to verify these formal abstraction and obtaining bounds on the actual policy.

\paragraph{Related work.}
The recently developed MoGym~\cite{grosmogym} is built on top of the~MODEST~toolset~\cite{DBLP:conf/tacas/HartmannsH14}, and it is related to our tool. 
The main difference is that our tool supports so-called permissive policies and feature remappings.
Likewise, Bacci et al.\ tackle the problem of verifying stochastic RL policies for continuous environments via an abstraction approach
\cite{DBLP:journals/corr/abs-2201-03698} but they abstract the policy using mixed-integer LPs. This limits the NN architectures their tool can handle. In contrast, we take the trained policy exactly as is and are architecture-agnostic.
Gu et al. propose a method called MCRL, that combines
model checking with reinforcement learning in a multi-agent setting for mission planning to ensure that safety-critical requirements are met \cite{DBLP:journals/sttt/GuJPSEL22}.
The difference in our approach is that we allow the post-verification of single RL policies in every kind of environment that can be modeled in the PRISM language.
Mungojerrie uses RL to obtain an optimal policy w.r.t. a given $\omega$-regular objective~\cite{DBLP:journals/corr/abs-2106-09161}.
In~\cite{DBLP:conf/formats/HasanbeigKA20}, a similar approach is presented for objectives given in linear temporal logic (LTL). 
Shielded RL guides the RL agent during training to avoid safety violations \cite{DBLP:conf/aaai/AlshiekhBEKNT18,jansen2020safe,DBLP:conf/tacas/DavidJLMT15,DBLP:conf/atva/DavidJLLLST14}. 
We, on the other hand, do not guide the training process but verify the trained~policy.

\section{Core Functionality and Architecture of COOL-MC}\label{sec:func_arch}

We now present the core functionality and the architecture of COOL-MC.

\paragraph{Training.}\label{sec:model_building}
During RL training, interaction with the user-provided environment takes place as follows. 
Starting from the current state, the agent selects an action upon which the environment returns the next state and the corresponding reward to derive a near-optimal policy.

\paragraph{Verification.}\label{sec:verification}
To allow model checking the trained policy (an RL agent) against a user-provided specification and formal model, we build the induced DTMC on the fly via Storm callback functions.
For every reachable state by the policy,
the policy is queried for an action.
In the underlying MDP, only states that may be reached via that action are expanded.
The resulting model is fully probabilistic, as no action choices are left open. It is, in fact, the Markov chain induced by the original MDP and the policy.

\vspace{-0.2em}
\paragraph{Policy transformations.}\label{sec:ab}
We extend the above-described on-the-fly construction with two abstractions.
One of them relies on feature abstraction \cite{DBLP:journals/corr/DragerFK0U15,DBLP:conf/tacas/Junges0DTK16}, while the other uses a coarser abstraction of the state variables  \cite{DBLP:conf/atva/JaegerJLLST19}. In both cases, we obtain a new policy $\tau$.

    {\textbf{1. Feature abstraction:}} Assume the state space has some structure $S \subseteq Q \times I$ for suitable $Q$ and $I$. Thus,  states are pairs $s = (q,i)$ where $q$ and $i$ correspond to \emph{features} whose values range over $Q$ and $I$ respectively. Let $\textsf{Act}$ denote the actions in the MDP.
    Given a partition $K_1,\dots,K_n$ of $I$, we abstract this feature from the policy $\pi$, by defining a \emph{permissive} policy~\cite{DBLP:journals/corr/DragerFK0U15} $\tau \colon S \rightarrow 2^\textsf{Act}$, i.e., a policy selecting multiple actions in every state. In particular, we consider
    $\tau(q,i) = \bigcup_{k \in K_n} \pi(q,k)$,
    with $K_n$ is the unique set such that $i \in K_n$.
This policy $\tau$ ignores the value of $i$ in state $(q,i)$ and instead selects any action that can be selected from states $(q,k)$, with $k \in K_n$.
    Applying a permissive policy yields an induced MDP, which can be model checked to provide best- and worst-case~bounds.

    \textbf{2. Feature remapping:} In this case, we again assume $S \subseteq Q \times I$. Given a mapping $\mu : I \to Y$ with $Y \subseteq I$, we define the abstracted policy $\tau$ as follows.
    $\tau(q,i) = \pi(q,\mu(i))$
    The feature values of $i$ are effectively being transformed into values from a (possibly smaller) set of feature values $Y \subseteq I$ before being fed into the original policy.



\begin{figure}[t]
    \centering
    \begin{subfigure}[b]{0.4\textwidth}
        \begin{tikzpicture}[
std/.style={
  draw,
  text width=2.5cm,
  align=center,
  font=\strut\sffamily
  },
rnd/.style={
  draw=#1,
  rounded corners=8pt,
  line width=1pt,
  align=top,
  text width=3cm,
  minimum height=2cm,
  font=\strut\sffamily
  },
bar/.style={
  <->,
  >=latex
  },
vac/.style={
  text width=2.5cm,
  align=center,
  font=\strut\sffamily
  },
ar/.style={
  ->,
  >=latex
  },
node distance=0.5cm and 3cm    
]

\node[vac] (rl_agent1)
  {RL Agent$_1$};
\node[vac,right=0.3cm of rl_agent1] (rl_agent2)
  {RL Agent$_n$};

\node[vac,text width=3.2cm,below=0.3cm of rl_agent1] (env1)
  {MDP-Environment$_1$};
\node[vac,below=0.3cm of rl_agent2] (env2)
  {Environment$_n$};
  
\node[draw,solid,inner sep=0pt,fit={(env1) (env2) (rl_agent1) (rl_agent2)}]
  (fit) {};

\draw[ar]
  ([xshift=-5pt]rl_agent1.south) -- ([xshift=-5pt]env1.north);
\draw[ar]
  ([xshift=5pt]env1.north) -- ([xshift=5pt]rl_agent1.south);

\draw[ar]
  ([xshift=-5pt]rl_agent2.south) -- ([xshift=-5pt]env2.north);
\draw[ar]
  ([xshift=5pt]env2.north) -- ([xshift=5pt]rl_agent2.south);

\draw [dashed] (-1.3,-1.58) -- (4.3,-1.58);

\node[text width=2cm] at (2.1,-0.5) {\emph{.......}};

\node[std,below=0.59cm of env1] (simulator)
  {Storm Simulator};
\node[std,below=0.59cm of env2] (openai){OpenAI Gym};

\draw[ar]
  ([xshift=-5pt,yshift=4.8pt]env1.south) -- ([xshift=-5pt]simulator.north);
\draw[ar]
  ([xshift=5pt]simulator.north) -- ([xshift=5pt,yshift=4.8pt]env1.south);  
  
\draw[ar]
  ([xshift=-5pt,yshift=4.8pt]env2.south) -- ([xshift=-5pt]openai.north);
\draw[ar]
  ([xshift=5pt]openai.north) -- ([xshift=5pt,yshift=4.8pt]env2.south);  

\node[ font=\fontsize{2}{2}\selectfont,text width=3cm] at (2,-1.47) {\emph{OpenAI Gym Interface}};
\node[ font=\fontsize{2}{2}\selectfont,text width=3cm] at (2.6,0.17) {\emph{MLFlow}};

\end{tikzpicture}
        \caption{RL Training.}
        \label{rfidtest_xaxis}
    \end{subfigure}
    \begin{subfigure}[b]{0.52\textwidth}
        \begin{tikzpicture}[
std/.style={
  draw,
  text width=2.5cm,
  align=center,
  font=\strut\sffamily
  },
rnd/.style={
  draw=#1,
  rounded corners=8pt,
  line width=1pt,
  align=top,
  text width=3cm,
  minimum height=2cm,
  font=\strut\sffamily
  },
bar/.style={
  <->,
  >=latex
  },
vac/.style={
  text width=2.5cm,
  align=center,
  font=\strut\sffamily
  },
ar/.style={
  ->,
  >=latex
  },
node distance=0.5cm and 3cm    
]

\node[vac,align=left] (experiment)
  {MLFlow};
\node[std,right=0cm of experiment,align=left] (environment)
  {Environment};
\node[std,right=0.2cm of environment,align=left] (rl_agent)
  {RL Agent};
\node[draw,solid,inner sep=5pt,fit={(experiment) (rl_agent)}]
  (fit) {};

\node[vac,align=left,below=0.5cm of experiment] (model_checking_learning)
  {New!};
\node[std,right=0cm of model_checking_learning,align=left] (model_builder)
  {Model Builder};
\node[right=2.7cm of model_builder,align=left] (openai_2gym)
  {};

\node[draw,dashed,inner sep=5pt,fit={(model_checking_learning) (openai_2gym)}]
  (fit3) {};

\node[vac,align=left,below=0.5cm of model_checking_learning,align=left] (storm_headline)
  {Storm};
\node[std,right=0cm of storm_headline,align=left] (simulator)
  {Simulator};
\node[std,right=0.2cm of simulator,align=left] (model_checker)
  {Model Checker};

\node[draw,solid,inner sep=5pt,fit={(storm_headline) (model_checker)}]
  (fit) {};


\draw[ar]
  ([yshift=0pt]environment.south) -- (model_builder.north);   

\draw[ar]
  ([yshift=0pt]simulator.north) -- (model_builder.south);   

\draw[ar]
  ([yshift=0pt]model_builder.east) -| (model_checker.north);   

\node[text width=2.2cm] at (5.7,-0.95) {\emph{Formal Model}};

\draw[ar]
  ([yshift=2pt]environment.east) -- ([yshift=2pt]rl_agent.west);   
 
\draw[ar]
  ([yshift=-2pt]rl_agent.west) -- ([yshift=-2pt]environment.east);   

\end{tikzpicture}
        \caption{RL Agent Verification.}
        \label{rfidtest_yaxis}
    \end{subfigure}
    \caption{COOL-MC Architecture.}
    \label{rfidtag_testing}
\end{figure}
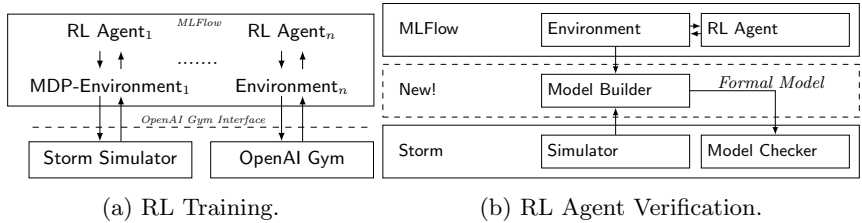

\paragraph{Architecture.}
COOL-MC is a Python toolchain on top of Storm and OpenAI gym. It also employs MLflow\footnote{MLflow is a platform to streamline ML development, including tracking experiments, packaging code into reproducible experiments, and sharing and deploying models~\cite{zaharia2018accelerating}.} as a file manager, which manages all the user experiments in the local file system, and allows the user to compare different experiments via the MLflow webserver. Furthermore, each task (for example, training or verification) is a separate MLflow component in the software architecture of COOL-MC and is controlled by the main script.
COOL-MC also contains the \emph{RL agent} component, which is a wrapper around the policy and interacts with the \emph{environment}.
Currently implemented agents include Q-Learning
\cite{watkins1992q}, Hillclimbing \cite{kimura1995reinforcement}, Deep
Q-Learning,
and REINFORCE~\cite{williams1992simple}.
For training uses the Storm simulator or an OpenAI Gym (Fig.
\ref{rfidtest_xaxis}). 
For verification, the model builder creates an induced DTMC which is then checked by Storm~(Fig.~\ref{rfidtest_yaxis}).

\section{Numerical Experiments}
Our experimental setup allows us to showcase the core features of COOL-MC.
For clarity of exposition,
in this paper, we focus on describing concise examples: 
In the \emph{Frozen lake}, the agent has to reach a Frisbee on a frozen lake. At every step, the agent can choose to move ``up’’, ``down’’, ``left’’, or ``right’’.
The execution of said movement is imprecise because of the ice: it is only as intended in $33.33\%$ of the cases. In the remaining $66.66\%$ of the cases, another movement is executed---the only restriction is that the agent cannot move in the direction opposite than its choice. The agent receives a reward of $+1$ if it reaches the Frisbee \cite{brockman2016openai}.
In the \emph{Taxi}
environment, the agent has to transport customers to their destination without running out of
fuel; in the \emph{Collision Avoidance} environment, it must avoid
two obstacles in a 2D grid; the descriptions of the other environments (\emph{Smart Grid, Stock Market, Atari James Bond, Atari Crazy Climber})\footnote{We refer the interested reader to the repository
  \url{https://github.com/DennisGross/COOL-MC} of the tool
  for more experiments with these and other environments.} can be found in the Appendix \ref{apx:environments}.
  We stress that the tool can handle several other benchmarks, e.g., PRISM-format MDPs from the \emph{Quantitative Verification Benchmark Set}
~\cite{DBLP:conf/tacas/HartmannsKPQR19}.
All experiments were executed on a 
laptop with 8~GB
RAM and an Intel(R)~Core(TM) i7-8750H CPU@2.20GHz $\times$ 12.

\begin{figure}[t]
    \centering
    \begin{subfigure}[b]{0.49\textwidth}
        \includestandalone[width=0.9\textwidth]{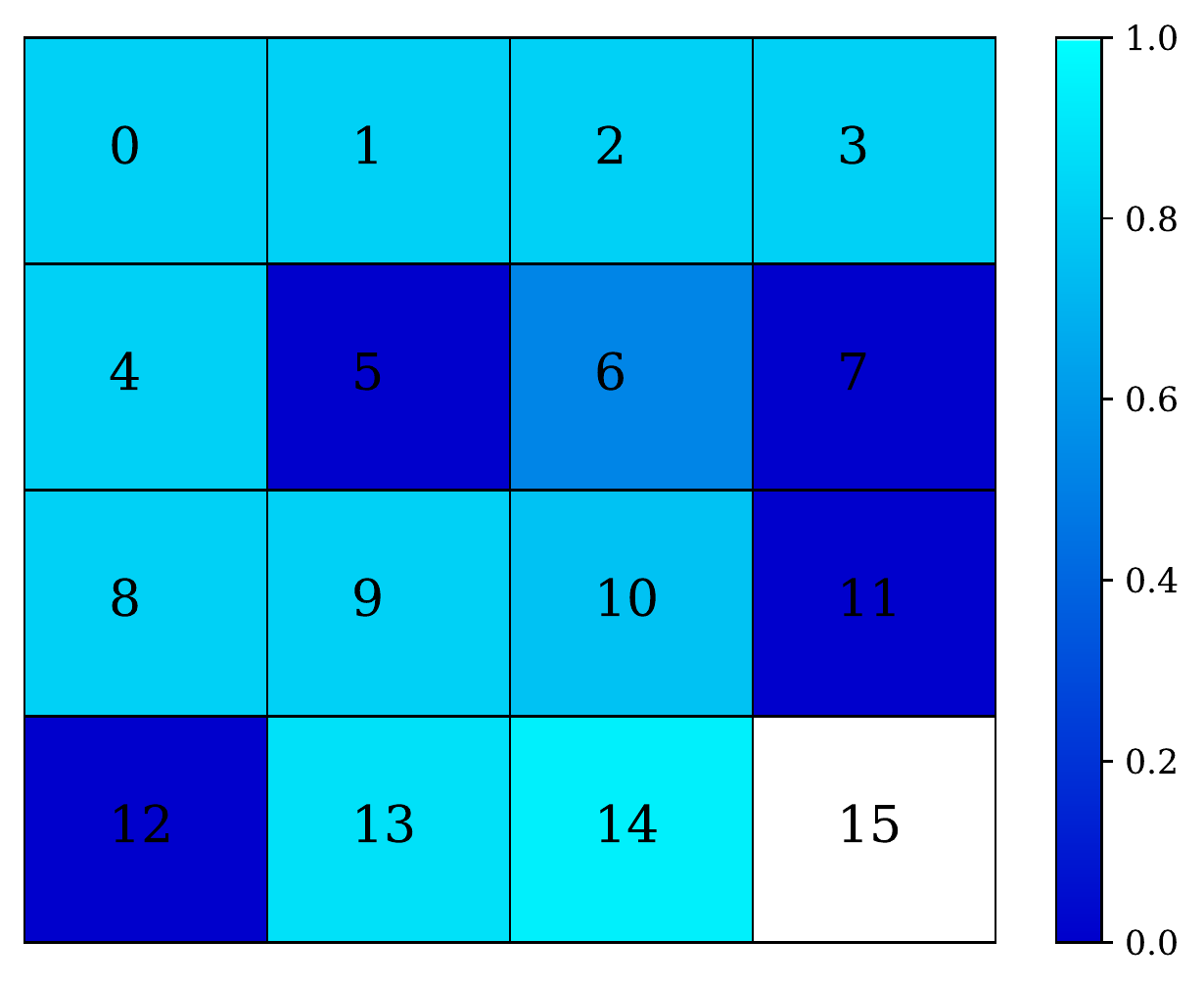}
        \caption{Heatmap shows per state $s$ \\the value $\mathcal{P}^{s,\pi_A}(\lozenge \mathit{frisbee})$.}
        \label{fig:frozen_lake}
    \end{subfigure}
    \begin{subfigure}[b]{0.5\textwidth}
        \includestandalone[width=1.05\textwidth]{fuel_precision}
        \caption{Plot shows different permissive taxi policies ``lumping'' fuel levels.}
        \label{fig:taxi_min_max}
    \end{subfigure}
    \caption{Frozen lake verification (a) and permissive taxi policies (b).}
    \label{rfidtag_testing}
\end{figure}

\begin{figure}[t]
    \centering
        \scalebox{0.82}{\input{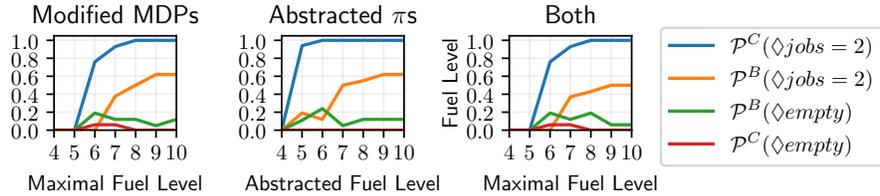}}
    \caption{Plots for the probabilities of running out of fuel on the road and finishing two jobs in three different settings for the trained policies $\pi_B$ and $\pi_C$.}
    \label{fig:abstract_fuel}
\end{figure}

\paragraph{Policy verification.}
The frozen lake environment is a commonly used OpenAI gym benchmark.
Therefore, we trained a deep-neural-network policy $\pi_{A}$ in this OpenAI gym for $100K$ episodes. For more information on the training process, see Appendix~\ref{apx:training}.
After the learning process, we verified $\pi_{A}$ in the frozen lake PRISM model (which describes the OpenAI environment). 
We investigate if the trained policy ultimately learned to account for the slippery factor.
Figure \ref{fig:frozen_lake} shows that the agent only falls into the water at the fifth position (from state $6$ while selecting the action ``left’’).
The agent reaches state $9$ with a probability of $\mathcal{P}^{0,\pi_A}(\lozenge 9)=1$. This indicates that the trained RL agent has learned to take the slippery factor into account since otherwise, it would not safely reach state $9$. At state $9$, we split the environment into an area in which the RL agent cannot fall into the water by following its policy and an unsafe area. If the agent successfully selects the action ``down'', it stays safe the rest of the trajectory to the Frisbee. Suppose the agent slips to state $10$ while selecting ``down'', the probability of reaching the Frisbee declines to $\mathcal{P}^{10,\pi_A}(\lozenge \mathit{frisbee}) = 0.76$. 

\paragraph{Scalability and Performance Analysis.}
A $10\times10$ instance of the
collision avoidance benchmark with a slickness constant value of
$0.1$ results in an MDP with $1,077,628$ states and $118,595,768$ transitions. This
already causes Storm to run out of memory. If we apply a policy
(trained with COOL-MC), the induced DTMC has $1M$ states and $29,256,683$
transitions, and is now within reach for Storm, while the result may not be
optimal. COOL-MC can handle sizes of up to $11 \times 11$ and a
slickness assignment of $0.1$ without running out of memory (see Table \ref{tab:benchmarks}). 
The bottleneck of our tool is the model building time (see Table \ref{tab:benchmarks}).
Model checking times are negligible in comparison.

\paragraph{Feature remapping.} 
The policies $\pi_B$ and $\pi_C$ are trained in the taxi environment to transport passengers to their destinations.
We are now interested in what happens if the taxi policies are deployed in cars with different maximal fuel-tank capacities. We can measure how well they perform for different such capacities using COOL-MC (left plot, Fig.~\ref{fig:abstract_fuel}).
On the other hand, to reduce design or hardware costs, one might be interested in deploying a single policy with a ``virtual'' or abstracted maximal fuel-tank capacity. Feature remapping can obtain such policies from $\pi_B$ or $\pi_C$. For example, an abstract fuel level of $6$ means that all fuel levels $6$--$10$ will be perceived as fuel level $6$. The modified policy can then be evaluated in the original MDP (middle plot, Fig.~\ref{fig:abstract_fuel}). Perhaps more interestingly, one can choose a particular abstraction and evaluate its performance at different physical max fuel-tank capacities (right plot, Fig.~\ref{fig:abstract_fuel}).

\paragraph{Feature abstraction.}
In Figure \ref{fig:taxi_min_max}, we first transform the trained policies $\pi_B,\pi_C$ into permissive ones $\tau_B,\tau_C$ due to, for instance, the lack of exact fuel-level sensors in the deployment hardware. To counter this loss of precision, we can get a best- and worst-case analysis under $\tau_B,\tau_C$ for different fuel-level ranges.
For example, a starting range of 8 means that COOL-MC creates a permissive policy that ``lumps'' fuel-levels 8, 9, and 10 together. Permissive policies with larger starting ranges have min/max probabilities closer together. 
\renewcommand{\arraystretch}{1.2}
\begin{table}[t]
\scalebox{0.78}{
\begin{tabular}{|l|lr|lllllr|}
\hline
\textbf{}            & \multicolumn{2}{l|}{\textbf{MDP}}                           & \multicolumn{6}{l|}{\textbf{DTMC}}                                                                                                                                                                                                                                             \\ \hline
\textbf{Environment} & \multicolumn{1}{l|}{\textbf{States}} & \multicolumn{1}{|l|}{\textbf{Trans.}} &
\multicolumn{1}{l|}{\textbf{Specification}}                                   & \multicolumn{1}{l|}{\textbf{Result}} & \multicolumn{1}{l|}{\textbf{Build Time}} & \multicolumn{1}{l|}{\textbf{Check. Time}} & \multicolumn{1}{l|}{\textbf{States}} & \multicolumn{1}{|l|}{\textbf{Trans.}}  \\ \hline
Frozen Lake          & \multicolumn{1}{r|}{17}              & 152                  & \multicolumn{1}{l|}{$\mathcal{P}^{\pi_A}(\lozenge \mathit{water})$}                               & \multicolumn{1}{r|}{0.18}            & \multicolumn{1}{r|}{0.25}                   & \multicolumn{1}{r|}{0}                      & \multicolumn{1}{r|}{14}              & 34                   \\ \hline
Taxi                 & \multicolumn{1}{r|}{8576}            & 39608                & \multicolumn{1}{l|}{$\mathcal{P}^{\pi_C}(\lozenge \mathit{empty})$}      & \multicolumn{1}{r|}{0}               & \multicolumn{1}{r|}{4.78}                   & \multicolumn{1}{r|}{0}                      & \multicolumn{1}{r|}{252}             & 507                  \\ \hline
Taxi                 & \multicolumn{1}{r|}{8576}            & 39608                & \multicolumn{1}{l|}{$\mathcal{P}^{\pi_C}(\lozenge 2)$}                                 & \multicolumn{1}{r|}{1}               & \multicolumn{1}{r|}{4.5}                    & \multicolumn{1}{r|}{0}                      & \multicolumn{1}{r|}{252}             & 507                  \\ \hline
Collision 10x10      & \multicolumn{1}{r|}{1077628}         & 118595768             & \multicolumn{1}{l|}{$\mathcal{T}^{\pi_D}(\lozenge \mathit{collide})$}                               & \multicolumn{1}{r|}{470.73}          & \multicolumn{1}{r|}{3106}                   & \multicolumn{1}{r|}{226}                    & \multicolumn{1}{r|}{1000000}         & 29256683             \\ \hline
Collision 11x11      & \multicolumn{1}{r|}{1885813}         & 211956692            & \multicolumn{1}{l|}{$\mathcal{T}^{\pi_D}(\lozenge \mathit{collide})$}                               & \multicolumn{1}{r|}{546.18}          & \multicolumn{1}{r|}{3909}                   & \multicolumn{1}{r|}{314}                    & \multicolumn{1}{r|}{1771561}         & 52433826             \\ \hline
Collision 12x12      & \multicolumn{1}{r|}{3148444}         & 359797152            & \multicolumn{1}{l|}{$\mathcal{T}^{\pi_D}(\lozenge \mathit{collide})$}                               & \multicolumn{1}{r|}{FAILED}          & \multicolumn{1}{r|}{6619}                   & \multicolumn{1}{r|}{FAILED}                 & \multicolumn{1}{r|}{2985984}         & 89198366             \\ \hline
Smart Grid           & \multicolumn{1}{r|}{271}             & 10144                & \multicolumn{1}{l|}{$\mathcal{P}^{\pi_E}_{\leq 1000}(\lozenge \mathit{black})$}            & \multicolumn{1}{r|}{0.02}            & \multicolumn{1}{r|}{0.48}                   & \multicolumn{1}{r|}{0}                      & \multicolumn{1}{r|}{40}              & 176                  \\ \hline
Stock Market         & \multicolumn{1}{r|}{14760}           & 89506                & \multicolumn{1}{l|}{$\mathcal{P}^{\pi_F}_{}(\lozenge \mathit{loss})$} & \multicolumn{1}{r|}{0}               & \multicolumn{1}{r|}{0.3}                    & \multicolumn{1}{r|}{0}                      & \multicolumn{1}{r|}{130}             & 377                  \\ \hline
James Bond           & \multicolumn{1}{r|}{172032}          & 3182592              & \multicolumn{1}{l|}{$\mathcal{P}^{\pi_G}_{\leq 15}(\lozenge \mathit{done})$}                      & \multicolumn{1}{r|}{0.23}            & \multicolumn{1}{r|}{1997}                   & \multicolumn{1}{r|}{0.28}                   & \multicolumn{1}{r|}{159744}          & 1105920              \\ \hline
Crazy Climber        & \multicolumn{1}{r|}{57344}           & 499712               & \multicolumn{1}{l|}{$\mathcal{P}^{\pi_G}_{\leq 15}(\lozenge \mathit{done})$}                      & \multicolumn{1}{r|}{0}               & \multicolumn{1}{r|}{175}                    & \multicolumn{1}{r|}{0}                      & \multicolumn{1}{r|}{8193}            & 32772                \\ \hline
\end{tabular}
}
\caption{Benchmarks for different environments and trained policies. Times are measured in seconds; \textit{FAILURE} stands for failure during model-checking.}
\label{tab:benchmarks}
\end{table}



\section{Conclusion and Future Work}
We presented the tool COOL-MC, which provides a tight interaction between model checking and reinforcement learning. 
In the future, we will extend the tool to directly incorporate safe reinforcement learning approaches~\cite{jansen2020safe,DBLP:conf/aiia/HasanbeigKA19,verifyinloop,DBLP:conf/cav/JothimuruganBBA22} and will extend the model expressivity to partially observable MDPs~\cite{DBLP:conf/cav/JungesJS20}.

\bibliographystyle{splncs04}
\bibliography{references}

\newpage
\appendix

\section{Environments}\label{apx:environments} 

\paragraph{Frozen Lake.} It is a commonly used OpenAI gym benchmark, where 
the agent has to reach the goal (frisbee) on a frozen lake. The movement direction of the agent is uncertain and only depends in $33.33\%$ of the cases on the chosen direction. In $66.66\%$ of the cases, the movement is noisy.
\begin{gather*}
States = \{0,1,2,3,4,5,6,7,8,9,10,11,12,13,14,15\} \\
Actions = \{up, left, down, right\}\\
Reward = \begin{cases}
        1 \text{, if state = 15 (Frisbee).}
        \\
        0 \text {, otherwise}.
        \end{cases}
\end{gather*}

\paragraph{Taxi.} The taxi agent has to pick up passengers and transport them to their destination without running out of fuel. The environment terminates as soon as the taxi agent does the predefined number of jobs. After the job is done, a new guest spawns randomly at one of the predefined locations.
If not further mentioned, we set the maximal taxi fuel level to ten and the maximal number of jobs to two.
\begin{gather*}
States =
\{(x,y,passenger\_loc\_x,passenger\_loc\_y,passenger\_dest\_x,\\passenger\_dest\_y,fuel,done,on\_board,jobs\_done,done),...\} \\
Actions = \{north,east,south,west,pick\_up,drop\}\\
Penalty = \begin{cases}
        0 \text{, if passenger successfully dropped.}
        \\
        21 \text {, if passenger got picked up.}
        \\
        21 + |x-passenger\_dest\_x| +\\ |y-passenger\_dest\_y| \text {, if passenger on board.}
        \\
        21 + |x-passenger\_loc\_x| + |y-passenger\_loc\_y| \text {, otherwise}
        \end{cases}
\end{gather*}


\paragraph{Collision Avoidance.} Collision avoidance is an environment that contains one agent and two moving obstacles in a two-dimensional grid world. The environment terminates as soon as a collision between the agent and one obstacle happens. The environment contains a slickness parameter, which defines the probability that the agent stays in the same cell.
\begin{gather*}
States = \{(x,y,obstacle\_x,obstacle\_y,obstacle\_x,obstacle\_y,done),...\} \\
Actions = \{north,east,south,west\}\\
Reward = \begin{cases}
        0 \text{, if collision}
        \\
        100 \text {, otherwise}
        \end{cases}
\end{gather*}

\paragraph{Smart Grid.} In this environment, a controller controls renewable- and non-renewable energy production distribution.
The objective is to minimize non-renewable energy production by using renewable technologies.
If the energy consumption exceeds production, it leads to a blackout. Furthermore, if there is too much energy in the electricity network, the energy production shuts down.
\begin{gather*}
States = \{(energy,blackout,renewable,non\_renewable,consumption),...\} \\
Actions = \{increase\_non\_renewable,increase\_non\_renewable,\\decrease\_renewable, decrease\_both\}\\
Penalty = \begin{cases}
        max(no\_renewable-renewable,0 )\text{, if no blackout.}
        \\
        1000 \text {, otherwise}
        \end{cases}
\end{gather*}

\paragraph{Stock Market.} This environment is a simplified version of a stock market simulation. The agent starts with an initial capital and has to increase it through buying and selling stocks without running into bankruptcy.

\begin{gather*}
States = \{(buy\_price, sell\_price, capital,stocks,\\last\_action\_price),...\} \\
Actions = \{buy, hold, sell\}\\
Reward = \begin{cases}
        \text{max(capital - initial capital, 0), if hold.}
        \\
        max(floor(\frac{capital}{buy\_price}),0)\text {, if buy.}
        \\
        max(capital+\text{number of stocks}\\\text{times sell\_price - initial capital, 0), if sell.}
        \\
        \end{cases}
\end{gather*}

\paragraph{James Bond.} This environment is an abstraction of the James Bond game for the Atari 2600. The goal is to collect rewards by shooting helicopters/diamonds and collecting diamonds. James Bond must avoid falling into radioactive pixels, which would terminate the environment.
The state space consists of images. Each image consists of three pixel rows with 6 pixels each (3x6), and one extra pixel as an auxiliary variable for PRISM.
The actions are \emph{jump}. \emph{tick}. and \emph{shoot}.
We refer to our repository for a detailed state space, action space, and reward~function.

\paragraph{Crazy Climber.}
The crazy climber is a game where the player has to climb a wall.
This environment is a PRISM abstraction based on this game.
Each state is an image.
A pixel with a One indicates the player position.
A pixel with a Zero indicates an empty pixel.
A pixel with a Three indicates a falling object.
A pixel with a four indicates the player's collision with an object.
The right side of the wall consists of a window front. The player has to avoid climbing up there since the windows are not stable.
For every level the play climbs, the player gets a reward of 1.
The player can also move left, right, or stay idle to avoid falling~obstacles.

\paragraph{Scalability.}\label{sec:scalability} Via the PRISM constant definitions, it is possible to scale up the smart grid, collision avoidance, stock market environment, and the Quantitative Verification Benchmark Set PRISM MDPs. The only limitation is the available~memory.

\section{Trained Policies}\label{apx:training}
We trained multiple policies for the previously introduced policies (see Table \ref{tab:other_trained_policies}).
We set for all experiments the following seeds:
\begin{itemize}
    \item Numpy Random Seed $= 128$
    \item PyTorch Random Seed $= 128$
    \item Storm Random Seed $= 128$
\end{itemize}
\begin{table}[t]
\scalebox{0.7}{
\begin{tabular}{|l|l|l|l|l|l|l|l|}
\hline
\textbf{Policy}  & \textbf{Environment}         & \textbf{Number of Layers} & \textbf{Number of Neurons} & \textbf{LR}     & \textbf{Batch Size} & \textbf{Episodes} & \textbf{Reward$_{100}$} \\ \hline
$\pi_A$ & Frozen Lake         & 4                & 128               & 0.001  & 128        & 100000    & 0.56          \\ \hline
$\pi_B$ & Taxi                & 4                & 512               & 0.0001 & 100        & 37000    & -1472.55         \\ \hline
$\pi_C$ & Taxi                & 4                & 512               & 0.0001 & 100        & 60000   & -1220.49      \\ \hline
$\pi_D$ & Collision Avoidance & 4                & 512               & 0.0001 & 100        & 49700   & 8995          \\ \hline
$\pi_E$ & Smart Grid & 4              & 512               & 0.0001 & 100        & 5000   & -1000          \\ \hline
$\pi_F$ & Stock Market & 3 & 128               & 0.0001 & 100        & 47612   & 371.07          \\ \hline
$\pi_G$ & James Bond & 4 & 512               & 0.0001 & 100        & 5000   & 13.43          \\ \hline
$\pi_H$ & Crazy Climber & 4 & 512               & 0.0001 & 100        & 3000   & 6         \\ \hline
\end{tabular}
}
\caption{All RL agents were trained with the standard deep Q-learning algorithm. 
With $\epsilon=1$, $\epsilon_{decay}=0.99999$, $\epsilon_{min}=0.1$, $\gamma=0.99$, and a target network replacement of 100. $Reward_{100}$ specifies the best reward sliding window over $100$ episodes during training.}
\label{tab:other_trained_policies}
\end{table}
For training, we set the following constant definitions for the PRISM environments:
The constant definition for the frozen lake environment is $slippery=0.333$.
The taxi environment defines $MAX\_FUEL=10,MAX\_JOBS=2$.
The collision avoidance environment defines $xMax=4,yMax=4,slickness=0$.
We define $max\_consumption=4,renewable\_limit=3,non\_renewable\_limit=3,grid\_upper\_bound=2$ for the smart grid environment.

\section{Probabilistic Computation Tree Logic (PCTL)}
We use the PCTL syntax in our experiments, which we now describe in more detail. Let $\lozenge E$ stand for the event that a state from the set $E$ is eventually reached. For MDPs, a (unique) probability measure is well-defined once we fix a deterministic policy $\pi$. Hence, we write $\mathcal{P}^{s,\pi}(\lozenge E)$ to denote the probability of the event $E$ if the initial state is $s$. The maximal probability value $\mathcal{P}_{max}^{s}(\lozenge E)$ is $\sup_\pi \mathcal{P}^{s,\pi}(\lozenge E)$, where the $\pi$ range over all deterministic policies; the minimal probability value $\mathcal{P}_{min}^{s}(\lozenge E)$ is defined similarly.

\end{document}